%% file: bare_jrnl_compsoc.tex
\documentclass[10pt,journal,compsoc]{IEEEtran}
\usepackage[sort,square,numbers]{natbib}

\usepackage{times}
\usepackage{epsfig}

\usepackage{subcaption}
\usepackage{placeins}
\usepackage{url}
\usepackage{booktabs}
\usepackage{amsmath}
\usepackage{amssymb}
\usepackage{graphicx}
\usepackage{xcolor}

\usepackage[linesnumbered,ruled]{algorithm2e}
\usepackage[utf8]{inputenc}
\usepackage[russian,english]{babel}

\usepackage{dblfloatfix}    

\graphicspath{ {images/} }

\begin{document}
%

\title{Rank-based verification for long-term face tracking in crowded scenes}

\author{Germán~Barquero, 
Isabelle~Hupont and Carles~Fernández 
\thanks{\indent Manuscript received December 30, 2020; revised April 13, 2021 and July 4, 2021; accepted July 20, 2021. This work was supported in part by
the Spanish Project AI-MARS (CIEN CDTI Programme) under Grant IDI-
20181108. This article was recommended for publication by Associate Editor
V. Štruc upon evaluation of the reviewers’ comments. (Corresponding author:
\indent Germán Barquero.)\newline \indent The authors are with the Research Department, Herta, 08037 Barcelona, Spain (e-mail: german.barquero@hertasecurity.com). \newline \indent Digital Object Identifier 10.1109/TBIOM.2021.3099568 \newline \newline \indent 2637-6407 © 2021 IEEE. Personal use is permitted, but republication/redistribution requires IEEE permission.\newline
See~https://www.ieee.org/publications/rights/index.html for more information.}}

\markboth{IEEE Transactions on Biometrics, Behavior, and Identity Science}%
{Shell \MakeLowercase{\textit{et al.}}: Bare Demo of IEEEtran.cls for Biometrics Council Journals}

\IEEEtitleabstractindextext{%
\begin{abstract}
Most current multi-object trackers focus on short-term tracking, and are based on deep and complex systems that often cannot operate in real-time, making them impractical for video-surveillance. In this paper we present a long-term, multi-face tracking architecture conceived for working in crowded contexts where faces are often the only visible part of a person. Our system benefits from advances in the fields of face detection and face recognition to achieve long-term tracking, and is particularly unconstrained to the motion and occlusions of people. It follows a tracking-by-detection approach, combining a fast short-term visual tracker with a novel online tracklet reconnection strategy grounded on rank-based face verification. The proposed rank-based constraint favours higher inter-class distance among tracklets, and reduces the propagation of errors due to wrong reconnections. Additionally, a correction module is included to correct past assignments with no extra computational cost. We present a series of experiments introducing novel specialized metrics for the evaluation of long-term tracking capabilities, and publicly release a video dataset with 10 manually annotated videos and a total length of 8' 54". Our findings validate the robustness of each of the proposed modules, and demonstrate that, in these challenging contexts, our approach yields up to 50\% longer tracks than state-of-the-art deep learning trackers.  
\end{abstract}

}

\maketitle

\IEEEdisplaynontitleabstractindextext

\IEEEpeerreviewmaketitle

\input{sections/01_introduction}
\input{sections/02_related_work}

\input{sections/03_methodology}

\input{sections/04_eval_metrics}

\input{sections/05_exp_results}
\input{sections/06_conclusions}

\ifCLASSOPTIONcompsoc
  \section*{Acknowledgments}
\else
  \section*{Acknowledgment}
\fi
This work was partly funded by the Spanish project AI-MARS (CIEN CDTI Programme, grant number IDI-20181108).

\bibliographystyle{unsrt}
\bibliography{IEEEabrv, bare_jrnl_compsoc}

\ifCLASSOPTIONcaptionsoff
  \newpage
\fi

%

\begin{IEEEbiography}[{\includegraphics[width=1in,height=1.3in,clip,keepaspectratio]{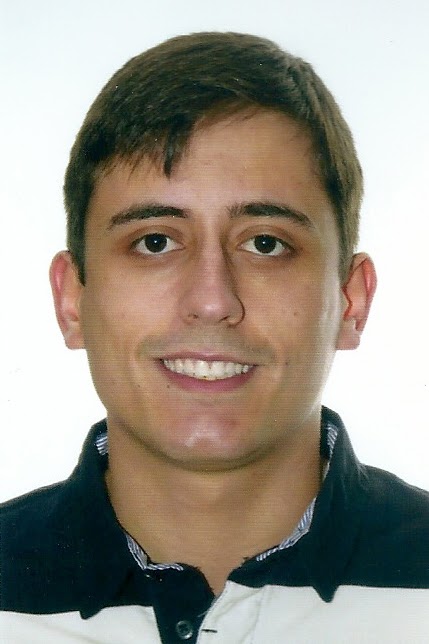}}]{German Barquero}
graduated from University of Barcelona (UB) with a bachelor's of science in Mathematics and Computer Science in 2020. He spent a year as a research assistant at École polytechnique fédérale de Lausanne (EPFL). He is currently enrolled in a MS program in Computer Vision at the Autonomous University of Barcelona (UAB). His research focuses on Computer Vision \& AI applied to medical imaging and biometrics.
\end{IEEEbiography}
\begin{IEEEbiography}[{\includegraphics[width=1in,height=1.3in,clip,keepaspectratio]{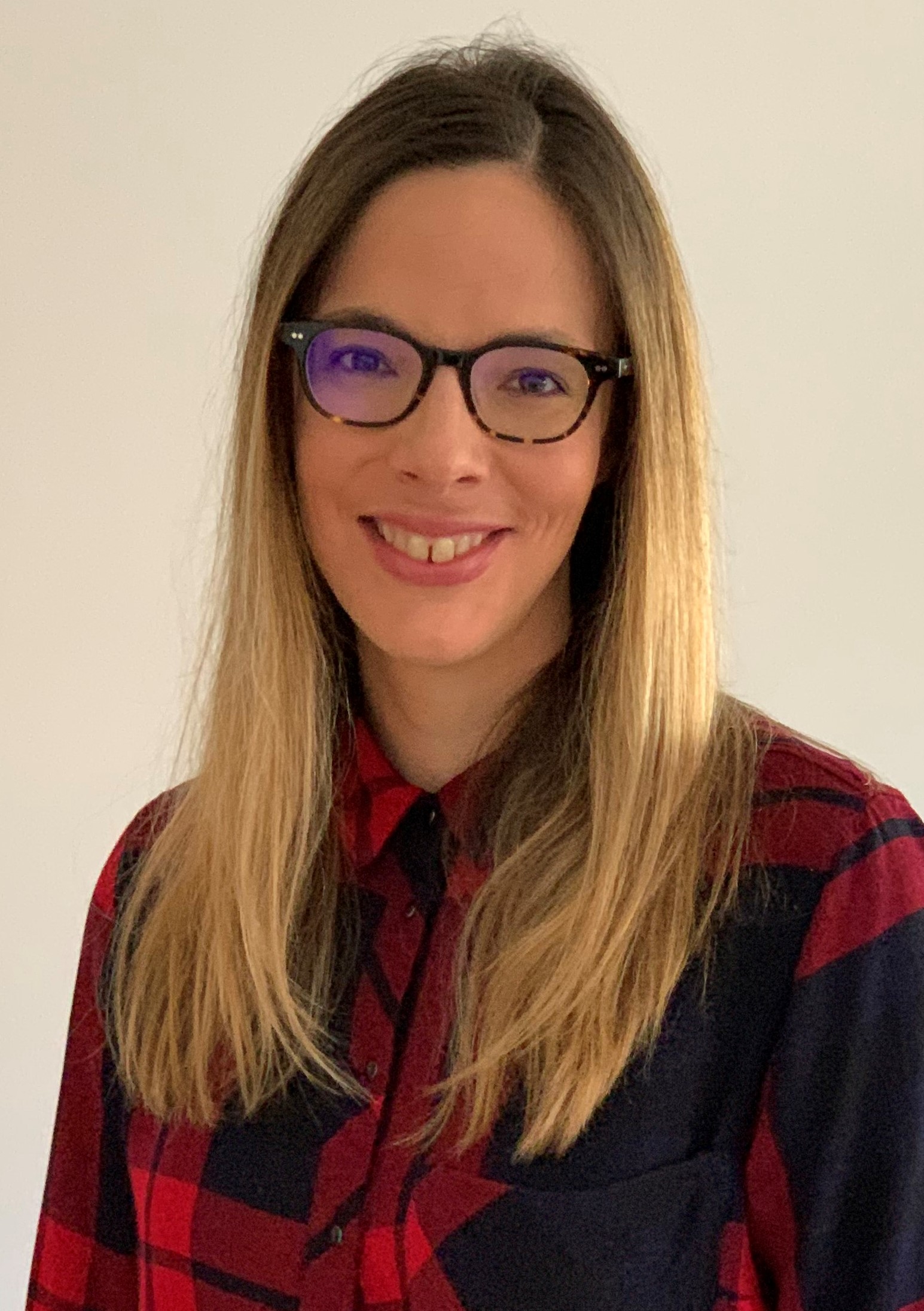}}]{Isabelle Hupont} received a PhD \textit{Cum Laude} in Computer Science in 2010 from the University of Zaragoza. From 2006 to 2015 she was a research manager at the Aragon Institute of Technology. She then spent two years as a senior researcher at ISIR - Sorbonne University, Paris. She is currently the Chief Research Officer at Herta, Barcelona. Her research focuses on artificial intelligence, affective computing, Human-Machine Interaction and face recognition. Isabelle has participated in more than 35 national and European public-funded R\&D projects, and has more than 50 international publications. 
\end{IEEEbiography}
\begin{IEEEbiography}[{\includegraphics[width=1in,height=1.3in,clip,keepaspectratio]{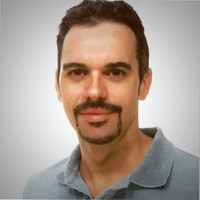}}]{Carles Fernández Tena} received a PhD Cum Laude in Computer Vision and AI from Universitat Autònoma de Barcelona in 2010, also receiving the 2010 Extraordinary PhD Award. He is currently CTO at Herta, where he has been leading the Research department since 2014. He has published more than 60 scientific articles in peer-reviewed international journals and conferences, and participated in public-funded projects from the FP6, FP7 and H2020 frameworks. His research interests include Deep Learning and Computer Vision, particularly face recognition and video analytics in very unconstrained environments.
\end{IEEEbiography}

\vfill

\end{document}

%% file: sections/01_introduction.tex
\IEEEraisesectionheading{\section{Introduction}\label{sec:introduction}}

\begin{figure}[t!]
\centering
\includegraphics[width=\linewidth]{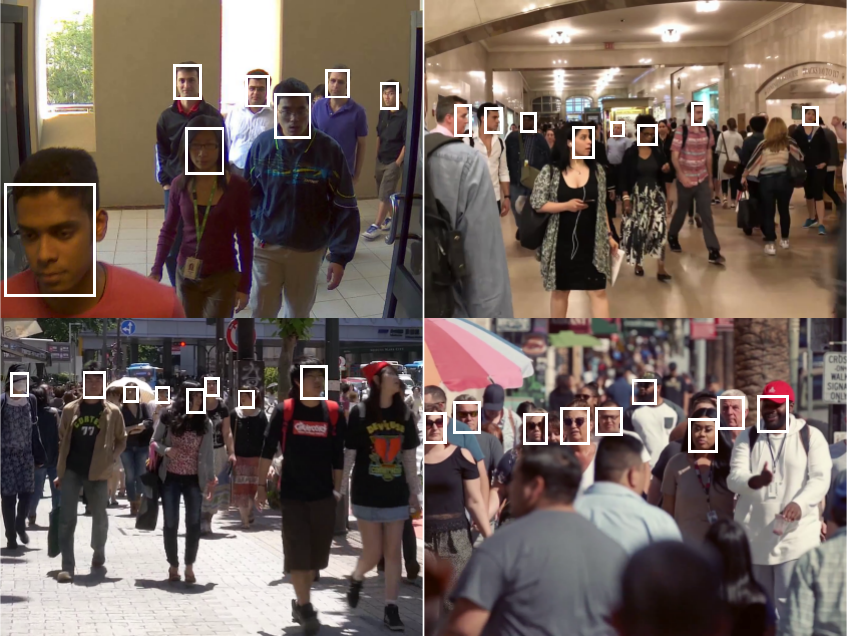}
\caption{Examples of near-eye level crowded video-surveillance scenes targeted in this work. Faces (white bounding boxes) are most of the time the only visible part of the subjects to be tracked. In these scenarios, the proposed face tracking approach surpasses full-body and general object tracking methods.}
\label{fig:crowded_onlyfaces}
\end{figure}

\IEEEPARstart{R}{ecent} advances on Convolutional Neural Networks (CNNs), IP cameras (optics, video compression, frame rates, ultra-high resolutions) and computational hardware (GPUs, DNN accelerators) have allowed video-surveillance systems to move to increasingly crowded, large-scale and unconstrained scenarios. These novel scenarios typically involve crowds of people massively walking toward cameras located at near-eye level, as illustrated in Figure~\ref{fig:crowded_onlyfaces}.


Such scenarios include large public open spaces (squares, large avenues, parks) and critical infrastructures (transport stations, airports, government buildings, malls) of the utmost importance for law enforcement bodies. There is a need for accurate video analytics systems in these locations, where major security threats occur (e.g. terrorist attacks) and where COVID-19 prevention measures have to be guaranteed.


Video-surveillance applications need to raise immediate alerts as a response to security and sanitary threats. For example, when a face recognition system is deployed, alarms are sent to end-users (e.g. police bodies) every time a new subject is detected or identified. However, the recurrent sending of duplicate alarms has to be controlled and minimized, in order not to collapse end-users. 
To avoid this infobesity problem, subjects need to be tracked, so that one single alarm is generated per track. Obtaining long and accurate tracks (e.g. without ID switches and fragmentations) is therefore essential to increase the system's usability. In the case of COVID-19 prevention, it is also extremely important to accurately track individuals at the long-term, so that people count is as exact as possible.

Real-time is also an important requirement for video analytics systems monitoring crowds. A subject may remain on scene for seconds or even minutes. Making enforcement bodies wait until the end of tracks to receive alerts (which is known as \textit{offline} strategy) implies losing a precious time that could save lives. Hence, online constraints become essential.

Nevertheless, from the algorithmic perspective, it is particularly challenging to perform reliable online and long-term tracking of subjects in this context since: 
\begin{itemize}
    \item The high pedestrian densities observed in crowded video-surveilled places lead to reduced pedestrian visibility. This results in frequent mutual occlusions which impede the full-body detection, deteriorating the performance of today's most common detection-based tracking algorithms \cite{2021trackingheads}. Consequently, tracking algorithms have to rely exclusively on the facial region ---which is most of the time the only visible part of the subjects--- instead of full-body or upper-body regions as in classic pedestrian tracking works \cite{li2014datasetCUHK,chen2019integrated}.
    \item People in video-surveilled places typically move all around the scene, positioning themselves closer or farther from the camera focus, and becoming again occluded or blurred for long periods. Existing generic object trackers cannot handle these situations properly, as demonstrated in \cite{lin2019mobiface}.
    \item Current state-of-the-art trackers, especially those based on Deep Learning, have a high computational cost \cite{CIAPARRONE202061}. While their real-time deployment may be possible in low-to-moderate pedestrian density scenes, computational performance drops dramatically in densely crowded scenarios.
    \item There is a lack of available datasets covering video-surveillance scenarios where persons' faces are recorded at near-eye level, so that they are visible enough to allow face detection and tracking at the microscopic level, but that show a macroscopic view of the crowd at the same time. Crowded video-surveillance videos recorded at near-eye level are particularly difficult to collect and annotate.
\end{itemize}

This work is an extension of \cite{barquero2020long}, in which we proposed an architecture especially conceived for long-term face tracking in crowded video-surveillance contexts. More specifically, our main contributions can be summarized as follows:
\begin{itemize}
    \item Our architecture recovers from partial and full long-term occlusions thanks to a novel online tracklet reconnection module grounded on rank-based face verification techniques.
    \item We also propose a track correction module, which updates past track assignments with current information. This module has no extra computational cost, while considerably improving long-term tracking performance.
    \item We validate the system with regard to different state-of-the-art trackers, and present an ablation study quantifying the contribution of each proposed module. Four validation metrics designed to evaluate long-term tracking capabilities are introduced for that purpose.
    \item We publicly release to the community the video dataset used in our experiments, including ground truth track annotations. All our videos present novel crowd scenes recorded at near-eye level, where faces are visible enough to be analysed at the microscopic level, while also benefiting from a macroscopic view of the crowd.
\end{itemize}

In this extended version, we present an improved tracklet reconnection 
module that substantially outperforms the previous one in all validation videos~\cite{barquero2020long}. We extend the validation dataset with five new videos, which increase the total dataset length by 4' 4" (about twice the original duration) and represent more challenging scenarios in terms of illumination, people's motion (velocity, direction, etc.) and occlusions. We have also updated the analysis of the state-of-the-art, by including the most recent tracking methods. Two new trackers have been validated with our dataset and incorporated in our comparisons \cite{Bewley2016_sort, huang2020globaltrack}. Finally, we have included an experiment that replicates a typical video-surveillance context, in which long video sequences are simulated by means of distractors, to demonstrate the robustness of the proposed method in real-life settings.

%% file: sections/02_related_work.tex
\section{Related work}

\subsection{Multi-object tracking}

Multi-object tracking is commonly carried out by assigning a single-object tracker to  each target of interest. Current state-of-the-art single-object trackers are based on Deep Learning (DL). A typical DL approach consists in splitting the tracking process into two stages: first, a region proposal network extracts regions of interest from the image and, then, a discriminator selects the best region candidate for the target object. For the second stage, a number of approaches rely on Siamese CNNs \cite{zhu2018dasiamrpn, li2019siamrpn++, wang2019siammask, voigtlaender2019trackingsegm}. These trackers usually make strong temporal consistency assumptions to ensure certain computational efficiency. Alternatively, other works propose to break these constraints and enable the tracker to search for the target at arbitrary positions and scales \cite{huang2020globaltrack}. However, unconstrained trackers become unstable in scenarios where tracked objects are very similar.

To gain efficiency, single stage or ``one-shot'' trackers have been recently proposed \cite{voigtlaender2019mots, wang2019towards, zhang2020fairmot}. In contrast to two-stage methods, one-shot trackers exploit multi-task learning to perform both object detection and feature extraction using a single network. Despite the impressive trade-off between accuracy and computational performance, they have been only applied to pedestrian (full-body) tracking yet. Their translation to other use-cases, such as face-based tracking, still needs to be explored. Other works suggest saving a pool of templates from past track images, the most representative and different among them, and use them to match regions of interest \cite{sauer2019holistic}. Very recent works leverage temporal and spatial Transformers \cite{vaswani2017attention} to perform tracking-by-attention \cite{meinhardt2021trackformer, wu2021track, yan2021learning}. Although they are very promising, their performance in multi-object tracking and especially in face tracking still needs to be properly explored.

Nevertheless, a common drawback of DL-based methods is that they are computationally expensive and cannot handle long-term tracking just by themselves: they fail to re-locate out-of-view targets when re-appearing in the scene. We refer the reader to \cite{lin2019mobiface} for a comprehensive demonstration and to \cite{CIAPARRONE202061} for an in-depth survey on DL trackers. A longer-term DL tracker has been recently proposed in \cite{wangtracking}, but it requires an initial offline training of the model with images from the particular target to track, and thus it is not suitable for our use case.

More computationally efficient trackers are not based on DL, but also achieve competitive accuracies. A clear example is SORT \cite{Bewley2016_sort}, in which Kalman filters and the Hungarian algorithm are combined to associate detections. In \cite{bochinski2018viou}, the use of a Kernelized Correlation Filter (KCF) visual tracker is proposed to fill the gaps generated after applying a classic intersection-over-union (IOU) data association strategy between frame detections. Both algorithms work at high frame rates, but still rely heavily on IOU values, which makes them prone to ID-switches. Other high-performing visual trackers include: MOSSE~\cite{bolme2010visual} and CSRT~\cite{lukezic2017discriminative}, which are both based on correlation filters; and the Median Flow tracker \cite{kalal2010forward}, based on motion flow. However, again, these trackers tend to fail with long-term occlusions.

\begin{figure*}[ht!]
\centering
\includegraphics[width=\linewidth]{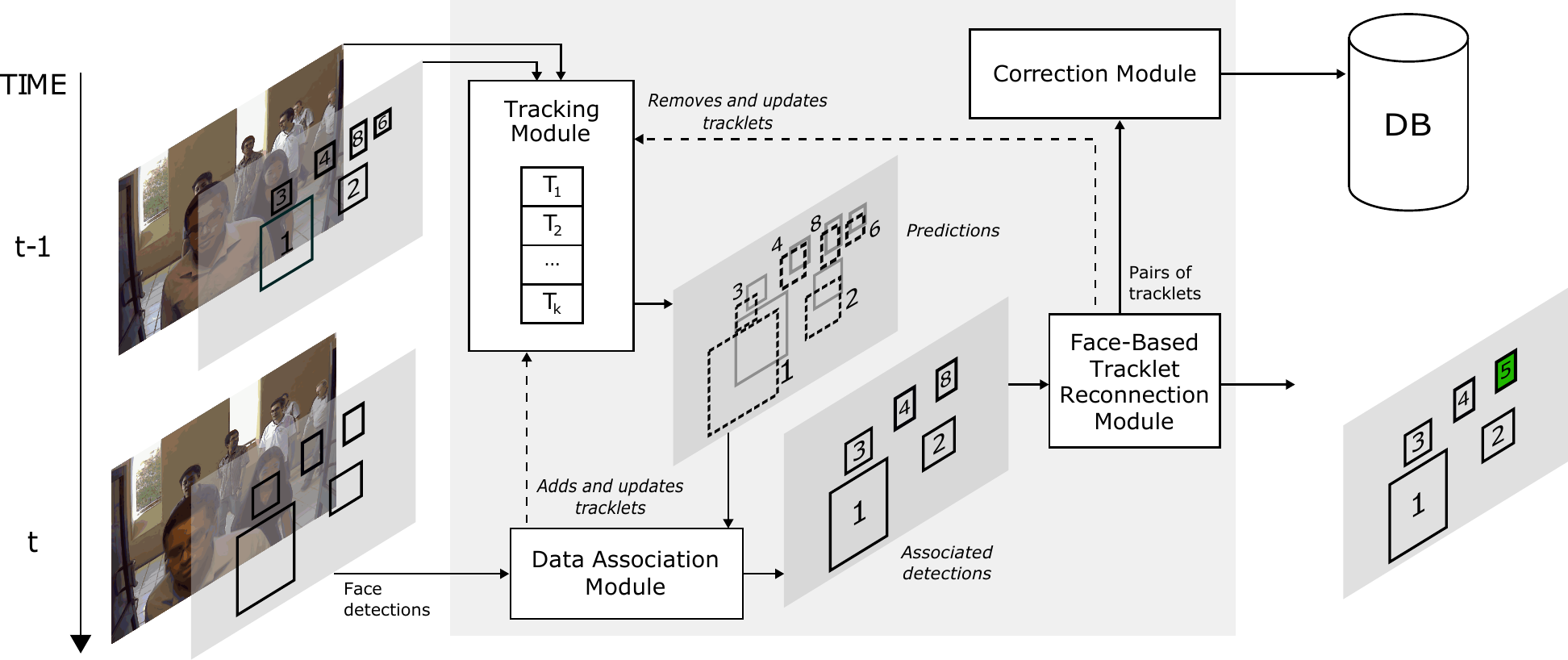}
\caption{Overview of the proposed four-module tracking architecture.}
\label{fig:system_scheme}
\end{figure*}

\subsection{Multi-face tracking}

The tracking of persons has been mainly tackled by applying generic object tracking approaches and targeting full-body regions. Few studies rely exclusively on faces to track persons and address face tracking as a problem with its own particularities. 

Taking advantage of the high accuracies reached by current face detectors, some face tracking works propose tracking-by-detection approaches. In \cite{comaschi2015online}, a generic AdaBoost face detector is combined with an adaptive structured output SVM tracker, using a IOU data association strategy. However, this approach is only suitable for short-term tracking, as it does not implement any tracklet reconnection strategy, and its core tracker \cite{hare2015struck} has been proven slower and less accurate than many newer trackers \cite{henriques2014comp2, bertinetto2016comp1, ou2018comp3}.
As a longer-term approach, \cite{kalal2010face} applies the Tracking-Learning-Detection (TLD) paradigm to faces: the face is tracked and simultaneously learned by a detector that supports the tracker once it fails. 

More recent approaches achieve long-term face tracking by using clustering techniques to associate short-term tracklets~\cite{lin2018offline, zhang2016offline}. Short-term tracklets are obtained by combining detectors and simple data association methods. Then, facial features are computed for each detection through DL face recognition models, and clusters are extracted from the feature space to collapse same-identity tracklets. Although these approaches achieve state-of-the-art results, they work fully offline and imply a high computational cost, which is not suitable for real-time tracking.

It is also worth mentioning that some works propose tracking mechanisms to improve face recognition in videos, in scenarios where persons of interest are previously enrolled using few still images. 
In \cite{dewan2016adaptive} and \cite{zheng2018automatic}, simple visual trackers are used to obtain tracks from which new (unseen) high-quality face stills are collected \cite{dewan2016adaptive}. These stills are matched against reference images to identify people. Interestingly, they are additionally used to enrich the gallery of enrolled images, thus improving face recognition performance. 
Nevertheless, these works focus on face recognition, leaving tracking as a secondary task.

\subsection{Datasets for face tracking in crowds}
\label{sec:public_datasets}

Studies on people tracking have traditionally focused on full-body pedestrian tracking in low-to-moderately crowded urban scenes. As a result, several pedestrian video datasets are available and commonly used by the community \cite{li2014datasetCUHK, liu2018datasetShanghaiTech, deng2014datasetPETA}. Another field for which a large number of datasets is available is crowd analysis, e.g. crowd counting or crowd behavior understanding~\cite{rodriguez2011datasetDriven, zhang2015datasetWorldExpo, dendorfer2019cvpr19}. These crowd datasets usually contain high-angle views, in which people faces appear at very low resolutions (mostly below 30x30 pixels). Some datasets have been conceived for face tracking and are thus closer to our scope. A relevant example is the dataset released in~\cite{zhang2016tracking}. However it focuses on sitcom and music videos taken with many different shots and not crowded at all (6 IDs per video). Also, the MobiFace dataset has been released to evaluate in-the-wild face tracking algorithms for mobile devices \cite{lin2019mobiface}. Videos are recorded from moving smartphone cameras, sometimes in ``selfie'' mode, and contain few faces (less than 5) per video. Consequently, none of these datasets cover our use case.

The only exception is the ChokePoint dataset. It provides a collection of 48 videos capturing individual subjects walking through two portals \cite{wong2011chokepoint}. To pose a more challenging real-world surveillance problem, two extra sequences were recorded in a indoor crowded environment, which represent the scenarios that we are targeting in this work.

%% file: sections/03_methodology.tex
\section{Proposed tracking architecture}

This work presents a four-module architecture that overcomes the limitations of previous approaches to favor long-term face tracking in crowded video-surveillance environments, see Figure \ref{fig:system_scheme}.   
The following sections describe each module in detail. 

\subsection{Tracking module (TM)}
\label{sec:STFT}

The system firstly extracts tracklets following a tracking-by-detection approach.
The tracking module is in charge of predicting face locations over the frames where the face detector fails, using one simple visual tracker. Several visual tracking algorithms are available for that purpose in our implementation, including KCF, MOSSE, Median Flow and CSRT.  

The tracking module creates a tracklet ${T_i}$ for every new detected face $i$. In case the detector loses a face in the following frame, it keeps predicting its position until the data association module decides to: (i) update the position with a new detection or (ii) force it to die. These tracklets are additionally used to collect the pool of reference images that serve as a basis for the face identification mechanism, c.f. Section \ref{sec:reID}.

\subsection{Data association module (DA)}
\label{sec:DA}

In order to decide which detection should guide which tracklet, a data association problem needs to be solved. 
Once the tracking module has predicted the new $K_t$ positions of live tracklets for a frame $t$, the data association module retrieves the $N_t$ faces detected in that frame. A state-of-the-art face detector is used in our implementation for that purpose \cite{zhang2017faceboxes}. Then, to establish correspondences between predicted and detected face bounding boxes, the Munkres implementation of the Hungarian algorithm is applied to their IOU values \cite{kuhn1955hungarian}. For every correspondence with an IOU value above a threshold $\lambda_{IOU}$, the tracking module updates the corresponding tracklet with the new detected bounding box position. For detected faces without a tracklet correspondence, a new tracklet is initialized and these faces are considered as new identities.

Tracklet predictions without a face detection correspondence are kept alive for $T_{max}$ frames. The tracking module keeps predicting their location over those frames where no association is made. If the tracklet is not updated for $T_{max}$ frames consecutively, it is forced to die and marked as inactive.

\subsection{Face-based tracklet reconnection module (FBTR)}
\label{sec:reID}

When a partial or full occlusion occurs, trackers generally lose the tracked target and consider it as a new object when it re-appears. To overcome this limitation, our system incorporates an online face-based tracklet reconnection module (FBTR). This module is inspired by face verification: a face recognition model is used to collect reference image templates from each tracklet, and then a matching procedure is applied to unify same-identity tracklets. In our implementation, we use the state-of-the-art face recognition model by \cite{deng2019arcface}.

\begin{figure}[t!]
\centering
\includegraphics[width=\linewidth]{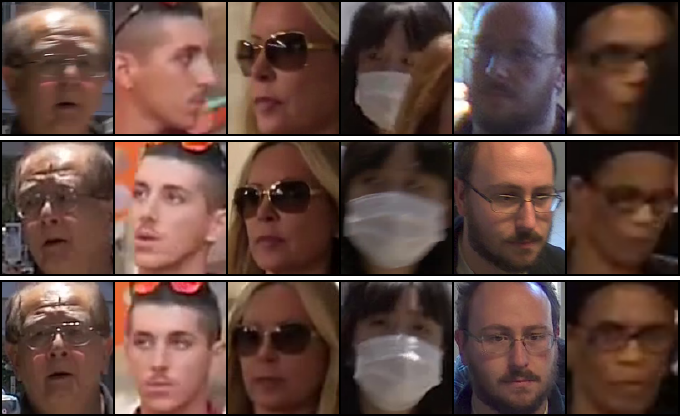}
\caption{Examples of discarded (top), verifiable (middle) and enrollable (bottom) faces.}
\label{fig:sample_snaps}
\end{figure}

The selection of reference templates is driven by image quality. More particularly, three indicators are considered: (i) face detection confidence, (ii) head pose angles and (iii) an image sharpness metric. Detection confidence is a value directly provided by the face detector \cite{zhang2017faceboxes}. Pitch, yaw and roll head pose angles are estimated using the fiducial extractor by Zhu et al. \cite{zhu20173ddfa}. Image sharpness is obtained by applying the Laplace operator in the facial area as proposed by Nikitin et al. \cite{nikitin2014facequality}. Using these quality indicators, face detections contained in each tracklet are divided into 3 groups (see Figure \ref{fig:sample_snaps}):

\begin{itemize}
\item \textit{Enrollable faces.} Faces with high visual quality, used to enroll identities.
\item \textit{Verifiable faces.} Faces that have enough quality to produce reliable templates. In the tracklet reconnection process, verifiable faces are matched against enrollable faces. Note that enrollable faces are a subset of verifiable faces.
\item \textit{Discarded faces.} Their low quality makes them unsuitable for the FBTR module, as they would produce non-reliable templates.
\end{itemize}


After data association, the FBTR component checks the quality of each detected face. If not discarded, a template is extracted with the face recognition model and stored either as enrollable or verifiable. Thus, in this work, the structure of a tracklet includes: (i) the set of detections associated to the same ID; and (ii) metadata in the form of image templates extracted by the face recognition model, either labelled as ``verifiable'' or ``enrollable''.

Then, for each tracklet $T_k\in\mathcal{T}$ with an assigned detection $D_k$ in the current frame, we retrieve tracklets $T_i\in\overline{\mathcal{T}}$, with $\overline{\mathcal{T}}=\mathcal{T}\setminus\{T_k\}$. For each retrieved tracklet $T_i$, the average of its enrollable face templates, $\overline{E_{T_i}}$, is computed and taken as the tracklet reference template. The average of all verifiable face templates of tracklet $T_k$, $\overline{V_{T_k}}$, is also computed. 

Now, letting $S$ be the similarity function of the face recognition model, we define $T^R$ as the rank-$R$ retrieved tracklet candidate, i.e. the tracklet at position $R$, after having sorted all candidates from highest to lowest similarity to $T_k$: 
\begin{equation}
    T^R=\underset{T_i\in \overline{\mathcal{T}}\setminus \{T^j\}_{1\leq j<R}}{\arg\max}
\hspace{1mm}S(\overline{E_{T_i}}, \overline{V_{T_k}})
\label{equation:topk}
\end{equation}

As part of the 
reconnection process, we fuse tracklet $T_k$ with $T^1$ (the rank-1 retrieved tracklet), but only when the following conditions are satisfied: 
\begin{enumerate}
\item The similarity between $T^1$ and $T_k$ is above a fixed threshold $\lambda_{FBTR}$:
\begin{equation}
S(\overline{E_{T^1}}, \overline{V_{T_k}})\geq \lambda_{FBTR}
\end{equation}
where $0\leq\lambda_{FBTR} \leq 1$.
\vspace{3mm}
\item That highest similarity is above the average of the next $C$-highest ones, by a margin $1/\epsilon$:
\begin{equation}
\label{equation:rank}
S(\overline{E_{T^1}}, \overline{V_{T_k}}) \geq \frac{1}{\epsilon} \cdot \frac{1}{C} \sum\limits_{r=2}^{C+1} S(\overline{E_{T^r}}, \overline{V_{T_k}})
\end{equation}
where $0<\epsilon\leq 1$ and $C\in\mathbb{N}{\geq 1}$.
\end{enumerate}

\noindent The rank-based constraint in Equation~\ref{equation:rank} helps filtering out wrong 
reconnections. This is particularly useful since, as the number and length of tracklets increase, tracking errors may result in mixed-up identity templates for a tracklet, which inevitably keeps on reducing inter-class distance among them. Thus, our second condition helps avoiding the propagation of reconnection
errors and allows us to be more permissible with the choice of $\lambda_{FBTR}$.


Therefore, whenever $T^1$ verifies the two previous conditions, the FBTR module re-assigns detection $D_k$ from tracklet $T_k$ to the most similar candidate $T^1$. Tracklets $T_k$ and $T^1$ are joined, and the pair $\langle T_k, T^1\rangle$ is appended to a list of track pairs, which is the input to the correction module.

\begin{figure}[t]
\centering
\includegraphics[width=0.87\linewidth]{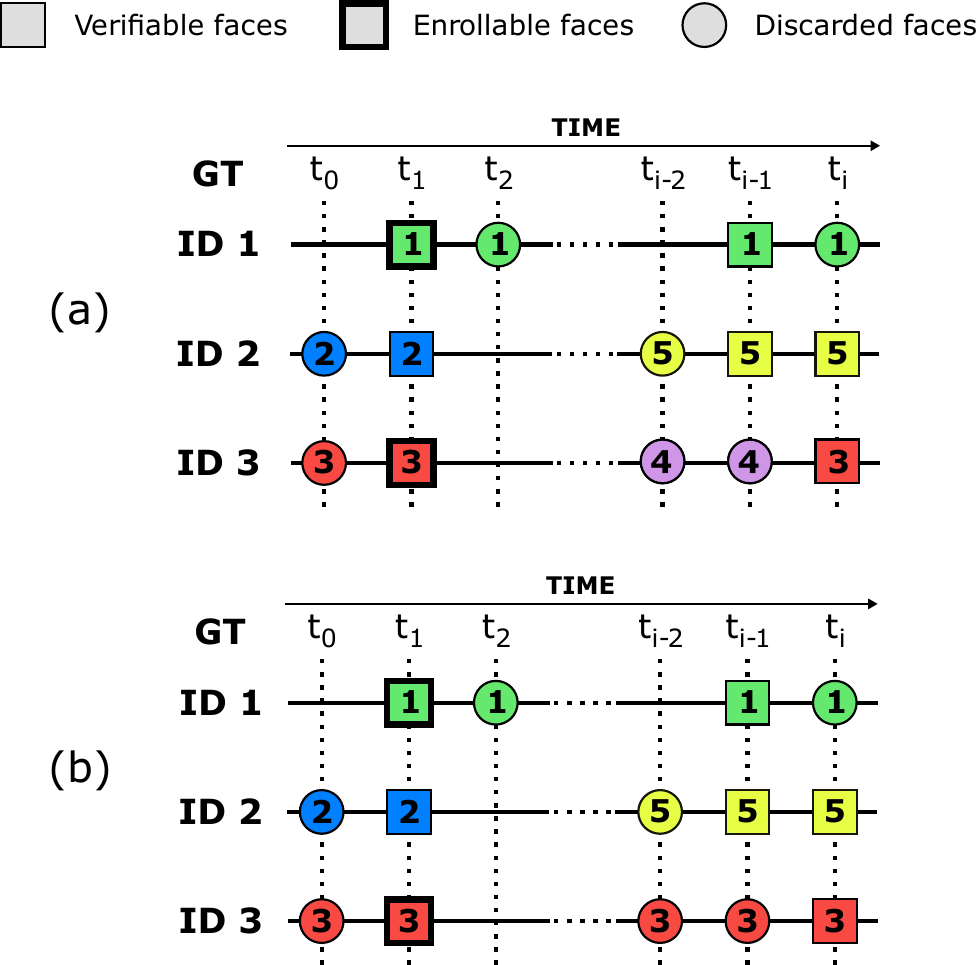}

\caption{Tracking results without~(a) and with~(b) correction module. Numbers inside detections correspond to the tracklet identifier assigned by our tracker. In (b), the reconnection of tracklet $T_4$ with $T_3$ at frame $t_{i}$ updates prior frames, too.}
\label{fig:reid_tracklets}
\end{figure}


To quantify the improvement brought by the rank-based reconnection criterion, both the current version of this module (FBTR) and the older one without the rank-based constraint (S\_FBTR)~\cite{barquero2020long} will be considered in our experiments.

\subsection{Correction module (CM)}
The correction module receives the pairs of tracklets joined by the FBTR module. For each pair 
$\langle T_k, T^1\rangle$, it retrieves all the detections assigned in the past to $T_k$ and switches their track ID to 
$T^1$. This strategy has a complexity of $O(n)$ (where $n$ is the number of detections inside the re-identified tracklet), so it can be implemented without adding significant computational cost, and is beneficial to refine the tracking history. This feature is particularly interesting for forensic video analysis.

Figure~\ref{fig:reid_tracklets} shows the effect of adding this module. The upper example (Fig.~\ref{fig:reid_tracklets}a) illustrates a regular tracklet reconnection: $T_4$ is reconnected to $T_3$ as soon as a verifiable face is found, whereas $T_5$ cannot be reconnected to $T_2$ as no enrolment face was available. On the other hand, the example at the bottom (Figure~\ref{fig:reid_tracklets}b) shows the effect of incorporating the correction module, which now replaces $T_4$ by $T_3$ not only for the current detection, but also the ones previous to instant $t_i$.

%% file: sections/04_eval_metrics.tex
\begin{table*}[ht!]
    \centering
    \small
    \begin{tabular}{ccccccccc}
        \toprule
         & Resolution & Length & Face dets. & Density & Scenario & \#Subjects & Description \\
         \midrule
        Choke1 & 800x600 & 1'24" & 7964 & 4.0 & Indoor & 24 (*) & Corridor recorded from 3 cameras over a door.\\
        Choke2 & 800x600 & 1'11" & 8710 & 4.8 & Indoor & 26 (*) & Corridor recorded from 3 cameras over a door.\\
        Terminal1 & 1920x1080 & 1'18" & 13722 & 5.9 & Indoor & 148 & A pedestrian scene filmed at eye level.\\
        Terminal2 & 1920x1080 & 1'15" & 11551 & 5.2 & Indoor & 140 & A pedestrian scene filmed at eye level.\\
        Terminal3 & 1920x1080 & 26" & 4255 & 5.5 & Indoor & 59 & A pedestrian scene filmed at eye level.\\
        Terminal4 & 1920x1080 & 35" & 6756 & 6.6 & Indoor & 126 & A pedestrian scene filmed at eye level.\vspace{1mm}\\
        Sidewalk & 1920x1080 & 27" & 8433 & 13.0 & Outdoor & 34 &  Crowd walking to the camera, filmed at eye level.\\
        Bengal & 1920x1080 & 40" & 6953 & 6.9 & Outdoor & 36 & A pedestrian scene filmed at eye level.\\
        Street & 1920x1080 & 1'8" & 4883 & 2.9 & Outdoor & 31 & Street scene filmed from a low-eye level angle.\\
        Shibuya & 3840x2160 & 30" & 8058 & 9.0 & Outdoor & 91 & A pedestrian scene filmed at eye level.\\
         \bottomrule
    \end{tabular}
    \caption{Description of the videos used to evaluate our tracking architecture. In videos marked with asterisk (*), subjects leave and re-enter the scene twice. Density refers to the mean number of face detections per frame.}
    \label{tab:videos}
\end{table*}

\section{Evaluation metrics}

In this section, we present novel metrics especially suitable for evaluating the long-term tracking performance of our system. Two of these metrics 
revisit and extend those commonly used in multi-object tracking, namely 
mismatch errors (number of ID switches produced across tracks) and track fragmentations~\cite{milan2016mot16}. The remaining two, Completion Rate Sum (CRS) and Completion Rate Plot (CRP), are introduced in this work for the first time.

One of the main challenges in multi-object tracking is to avoid drifting, i.e. losing a target. Assuming that the facial detector has high accuracy, in our use-case drifting only happens when two or more tracklets switch their target identity. The drifting effect becomes much more dramatic when incorporating reconnection capabilities: when switching its target, the tracklet integrity is lost, leaving the system vulnerable to future misassignments. Therefore, there is a need for revisiting the mismatch error metric by taking into account two new important concepts:

\begin{itemize}
    \item \textit{Soft-mismatch error (smme).} It is produced when the tracker switches the correct identity (ID) to a new one that has not been associated to any track until that time. This leads to ID-fragmentation, but can potentially be recovered by the FBTR module.
    \item \textit{Hard-mismatch error (hmme).} It is produced when, instead of switching to a new identity, the tracker switches to a previously assigned one. This leads to a probably unrecoverable ID-switch.
\end{itemize}

Another desired feature of our system is its ability to obtain long-term tracklets. As our track annotations only consider faces detected by the face detector, which were not corrected and therefore match the outputs of our architecture, traditional metrics like MOTA or MOTP \cite{bernardin2008evaluating} are not suitable for evaluating the tracking accuracy. Thus, it is necessary to introduce metrics able to quantify the length of tracklets generated by detection-based trackers. Overall, the goal is to build a robust long-term tracker that reduces fragmentation while keeping the number of ID-switches low. To achieve it, we formulate three new scalar metrics and a graphical one: \\

\noindent\textbf{Fragmentation (Frag)} 
\begin{equation*}
    Frag=\frac{\sum{smme}}{\#dets}
\end{equation*}
where the numerator is the sum of soft-mismatch errors produced in a video, and $\#dets$ is the total number of faces detected (according to ground truth annotations).\\

\noindent\textbf{ID-Switches (IDSW)}
\begin{equation*}
    IDSW=\frac{\sum{hmme}}{\#dets}
\end{equation*}
where $hmme$ is the total number of hard-mismatch errors in the video.\\

\noindent\textbf{Completion Rate Plot (CRP)}. 
Plot showing the percentage of tracks (vertical axis) that had at least $X\%$ of their ground truth detections correctly identified (horizontal axis). In other words, it plots $CR_{X}$ values for $0\leq X \leq 100$, where:
\vspace{1mm}
\begin{equation*}
    CR_X=\frac{\textit{\# IDs correctly tracked for at least X\% of the time}}{\textit{total number of IDs}}
\end{equation*}

\noindent\textbf{Completion Rate Sum (CRS)}. 
Area under the curve of the completion rate plot. The higher this value is, the longer subjects have been successfully tracked.
\begin{equation*}
    CRS=\frac{\sum_{X=1}^{100}{CR_X}}{100}\hspace{0.5cm}
\end{equation*}

%% file: sections/05_exp_results.tex

\section{Experiments and results}

This section describes the dataset used to evaluate the proposed architecture, and the results of the different experiments carried out with it. All experiments were conducted on a desktop Intel(R) Core(TM) i7-9700K CPU machine at 3.60GHz and a NVIDIA GeForce RTX 2070 GPU. The system thresholds $\lambda_{IOU}$, $\lambda_{FBTR}$ and $\lambda_{S\_FBTR}$, were adjusted using a private collection of training videos to 0.25, 0.5 and 0.7, respectively. The parameters of the FBTR module were set to $\epsilon=0.8$ and $C=6$, based on the experimental results presented in Section~\ref{section:FBTR_validation}. Quality thresholds for detection confidence, head angles and image sharpness were respectively set to 0.95, $\pm25^{\circ}$ and 0.9 for enrollable faces, and 0.8, $\pm60^\circ$ and 0.75 for verifiable faces.


\subsection{Evaluation dataset}

Since there are no public datasets fully corresponding to our use case (c.f. Section \ref{sec:public_datasets}), we have compiled and annotated a set of ten videos showing  crowded indoor and outdoor video-surveillance scenes. The total duration of the dataset is 8' 54''.

Two of these videos come from the extra sequences (cameras P2E\_S5 and P2L\_S5) of the well-known public dataset ChokePoint \cite{wong2011chokepoint}. To force re-appearances of subjects and validate the FBTR module performance, the three sequences recorded by P2E\_S5 cameras were concatenated, leading to the \textit{Choke1} video. Similarly, the video \textit{Choke2} was generated from \textit{P2L\_S5} cameras sequences. The remaining eight videos were selected from YouTube. Table \ref{tab:videos} details each video content and properties.

Tracks were semi-automatically annotated to obtain ground truth data. Firstly, face bounding boxes were retrieved using a state-of-the-art face detector \cite{zhang2017faceboxes}. Only faces with a detection confidence above 0.50 were considered. Then, detections were manually verified, and tracks and corresponding IDs were annotated\footnote{All videos and annotations are publicly available at \url{https://github.com/hertasecurity/LTFT}.}.

\subsection{Benchmarking of face trackers}

\begin{figure}[t!]
    \centering
    \includegraphics[width=0.95\linewidth]{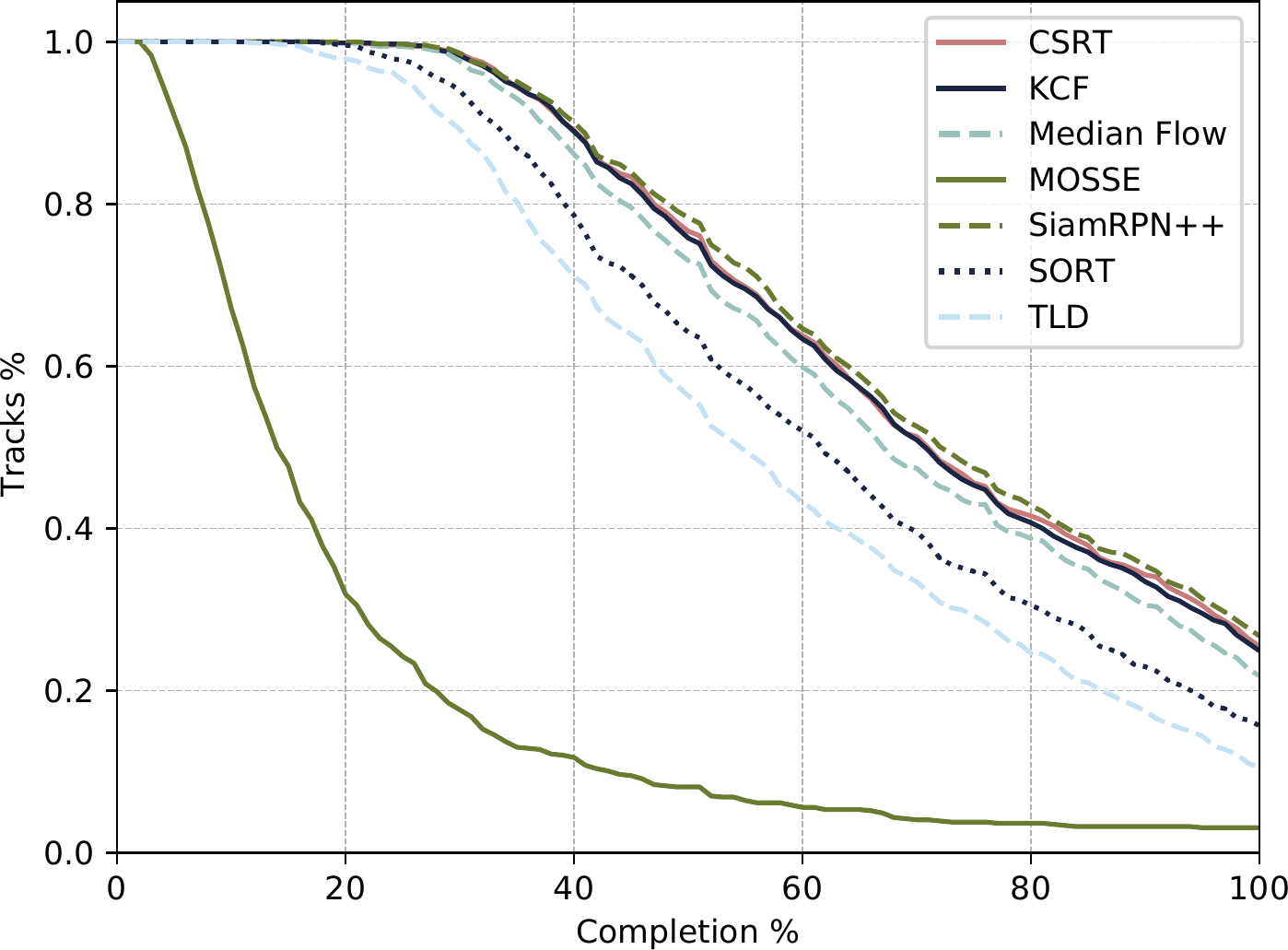}
    \caption{CRP for each tracker tested in the tracking module.}
    \label{fig:stft_CRP}
\end{figure}

\label{sec:STFT_benchmark}

This first experiment aims at benchmarking different trackers to choose the most appropriate one for the tracking module. We analyze the performance of our data association module using the following visual trackers: CSRT~\cite{lukezic2017discriminative}, KCF~\cite{bochinski2018viou}, Median Flow~\cite{kalal2010forward}, MOSSE~\cite{bolme2010visual} and SORT~\cite{Bewley2016_sort}. Additionally, we include in our benchmark the longer-term tracker TLD \cite{kalal2010face} and the state-of-the-art deep trackers GlobalTrack \cite{huang2020globaltrack} and SiamRPN++~\cite{li2019siamrpn++}. 

Table~\ref{tab:results_STFT} and Figure~\ref{fig:stft_CRP} present the results obtained on all of our evaluation videos. Results highlight the lower performance of TLD. This is probably due to its capability to re-identify targets, which makes it vulnerable to ID-switches. This effect is also observed with GlobalTrack. In this case, the deep learning models stored for each lost tracker quickly lead to a memory overflow, which prevents it from being evaluated under the same conditions. The best performing tracker in terms of ID-switches is MOSSE, but its high fragmentation leads to a poor overall completion rate. The SiamRPN++ tracker achieves the highest completion rate ($CRS=0.722$) and the lowest fragmentation ($Frag=0.01502$). However, KCF is very close to it ($CRS=0.710$, $Frag=0.01650$) and runs at much higher FPS, which is critical in video-surveillance contexts.
SORT provides the highest frame rate and a fairly low number of ID-Switches, in exchange of higher fragmentation and, consequently, a lower completion rate.

According to these findings, simple visual trackers and more sophisticated DL-based trackers lead to similar performances in such challenging environments where occlusions are extremely frequent. In the following experiments we will use the KCF tracker as the core of our tracking module.

\subsection{FBTR parameters}
\label{section:FBTR_validation}

In this section we analyze how the choice for parameters $\epsilon$ and $C$ affects the stability of the FBTR module. All the experiments in this section are run on a private collection of six videos, with a total length of 10' 57". 

To mimic the state of the FBTR module at any given point of the tracking timeline, we manually annotate the videos with ground truth detections and identity tracks, yielding a perfect tracking database with non-mixed identities. Note that such database combines identities from all video sequences. Furthermore, in order to simulate noisy tracking in real-life conditions, we generate additional databases with a predefined number of \textit{mixed identities}. This is achieved by randomly picking a pair of pure tracklets, and moving a slice of 5 consecutive templates from each of them to a new single tracklet, now containing the 5+5 mixed templates. By repeating this operation, we generate six databases in which an increasing percentage of identities are mixed (0, 5, 10, 15, 20 and 25\%).



%

Next, in order to validate the discriminating capabilities of the rank-based FBTR module depending on choice of $\epsilon$, we proceed as follows. For each identity in the database we randomly extract a slice of 5 verifiable templates, which simulates a \textit{target 
reconnection} when queried against the selected database. We additionally generate another query, which mixes 5+5 verifiable templates from two different identities, hence simulating a tracklet that underwent an \textit{ID-switch}. For each of these two queries, and values of $C=\{1\ldots 9\}$, we compute $\epsilon$ by isolating it in Equation~\ref{equation:rank}. In the case of the 
reconnection query, we consider either a correct or wrong 
reconnection, depending on whether the identity assigned by the highest similarity match is correct or not. This way, we obtain 3 distribution density curves (correct reconnections, 
wrong reconnections 
and ID-switches) for each combination of $C$ and dataset ($9\times 6=54$ experiments in total). Each process is repeated ten times to avoid edge results. 

\begin{table}[t!]
    \centering
    \small
    \begin{tabular}{ccccc}
        \toprule
        Tracker & Frag & ID-Switches & CRS & FPS\\
        \midrule
        CSRT & 0.01599 & 0.00459 & 0.714 & 5.0\\
        GlobalTrack & - & - & - & - \\
        KCF & 0.01650 & 0.00469 & 0.710 & 26.4\\
        Median Flow & 0.02243 & 0.00413 & 0.690 & 22.9\\
        MOSSE & 0.34383 & \textbf{0.00139} & 0.201 & 26.1\\
        SiamRPN++ & \textbf{0.01502} & 0.00565 & \textbf{0.722} & 2.1\\
        SORT* & 0.03151 & 0.00197 & 0.634 & \textbf{35.8}\\
        TLD & 0.06504 & 0.01127 & 0.585 & 1.6\\
        \bottomrule
    \end{tabular}
    \caption{Evaluation of the data association module on the whole dataset, using different trackers in the tracking module. (*) SORT was implemented with its default data association method.}
    \label{tab:results_STFT}
\end{table}

\begin{figure*}[t!]
    \centering
    \includegraphics[width=\textwidth]{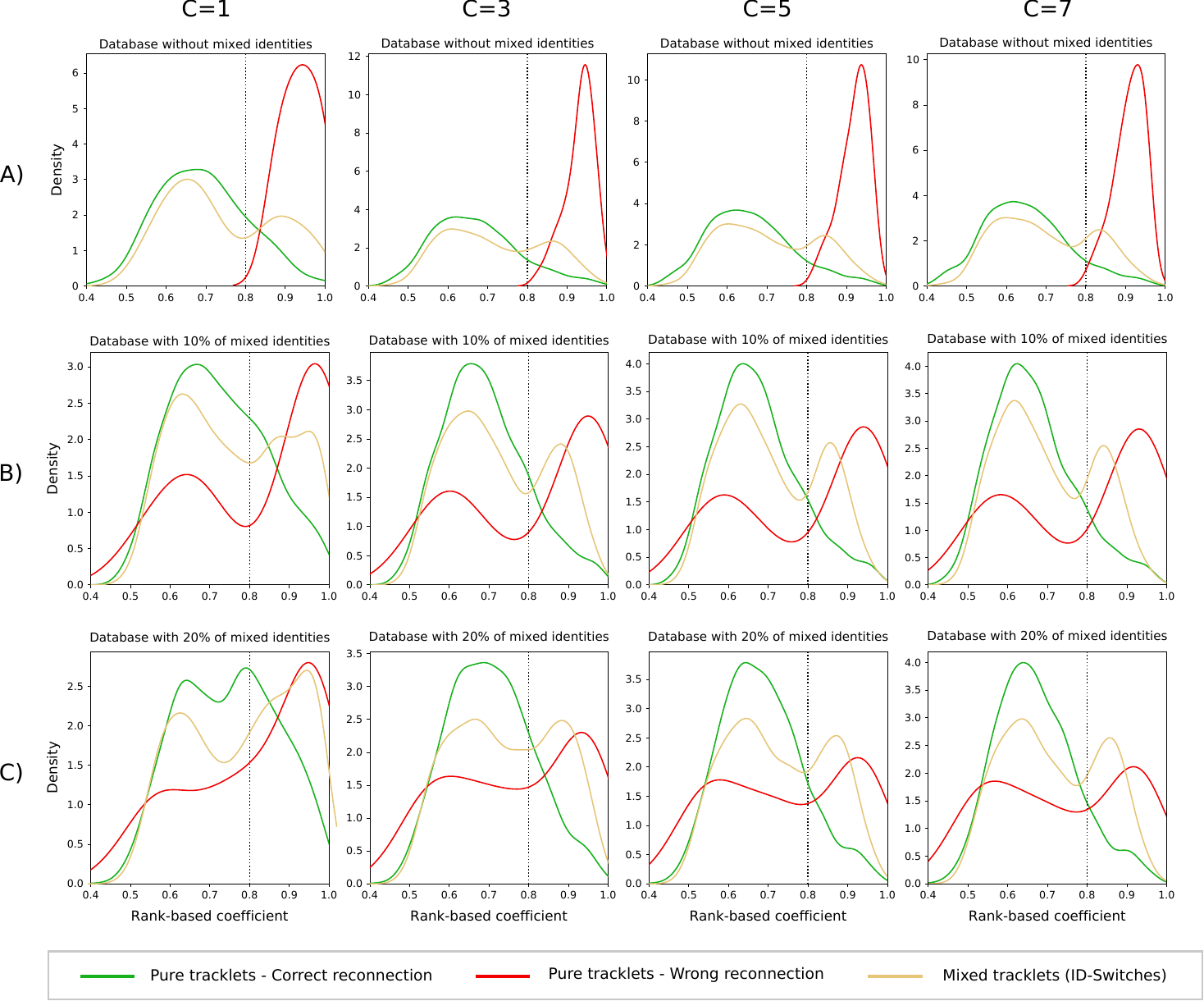}
    \caption{Distributions of the rank-based threshold $\epsilon$ for correct / wrong 
    reconnections (pure identities) and ID-switches (mixed identities). $C$ varies across columns. A), B) and C) use a FBTR database with 0\%, 10\% and 20\% of mixed identities, respectively.}
    \label{fig:exp1_epsilon}
\end{figure*}

\begin{figure*}[ht]
    \centering
    \includegraphics{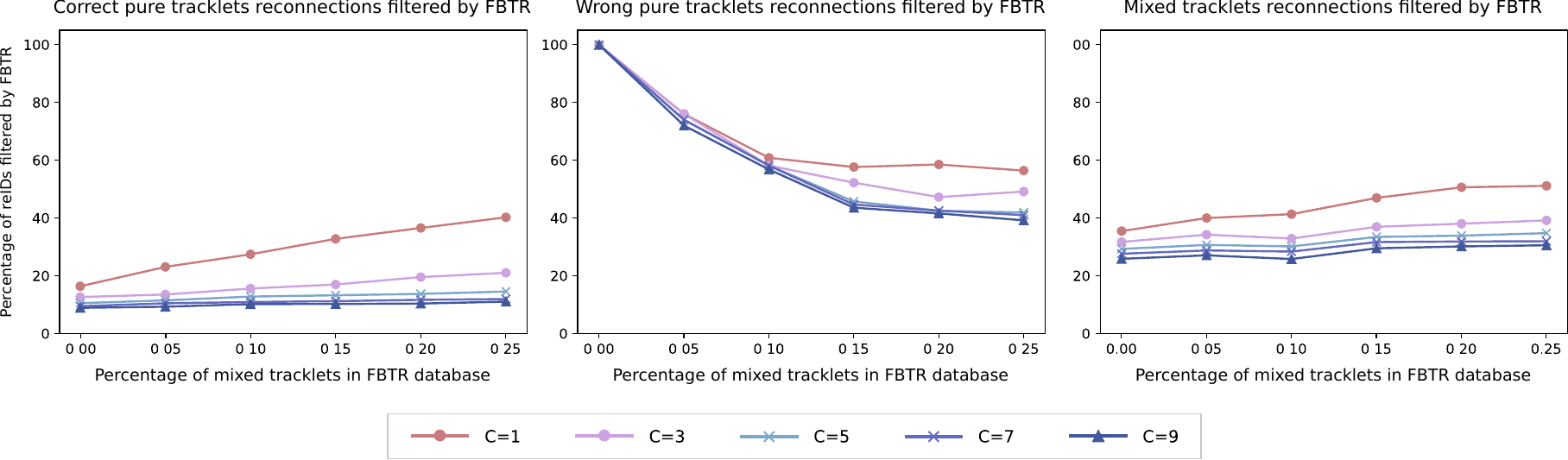}
    \caption{Percentage of filtered 
    reconnections for pure and mixed identities in function of $C$, with $\epsilon=0.8$.}
    \label{fig:exp2_c}
\end{figure*}


Figure~\ref{fig:exp1_epsilon} contains a subset of the resulting distributions, which show high discrimination between correct and wrong 
reconnections of pure identities, when no ID-switches are considered (row A). Although the  
reconnection of mixed identities poses a bigger challenge, a threshold $\epsilon=0.8$ seems to offer a good trade-off between correctly and wrongly filtered 
reconnections. This is desirable to ensure the stability of the FBTR module. When using only the best and second best similarity matches ($C=1$), the increase in mixed identities causes a dramatic distortion of the distributions, also decreasing the discriminating power of $\epsilon$. Favorably, $C$ acts as a distribution regularizer and helps reducing this effect. 

Additionally, Figure~\ref{fig:exp2_c} plots the percentage of filtered match candidates for $C\in\{1,3,5,7,9\}$ when setting $\epsilon=0.8$. As expected, we observe a significant improvement on high $C$ values as the proportion of mixed identities increases. For example, with 5\% of the identities mixed, any value of $C$ reduces around 75\% of wrong reconnections, 
but large values of $C$ reduce correct reconnections 
in only 10\%, as opposed to the undesirable reduction of over 20\% with $C=1$. A saturation plateau is reached around $C=6$, which is the setting used for the remaining experiments.

\subsection{Ablation study}
In this section, we present an ablation study that quantifies the contribution of each module (TM, DA, FBTR, and CM) and demonstrates the suitability of the proposed approach for long-term tracking in crowded video-surveillance scenarios. The following ablation experiments are presented: 

\begin{itemize}
    \item \textbf{DA.} Tracking is performed following a simple data association strategy: the tracking module is deactivated, and data association is computed based on the IOU value of detections in consecutive frames.
    \item \textbf{DA+TM.} In this experiment, data association is carried out using KFC in the tracking module. No face identification or correction mechanisms are applied.
    \item \textbf{DA+FBTR.} Same as DA, plus face identification. The two versions of the FBTR module are evaluated: the rank-based one (FBTR) and the simplified one (S\_FBTR).
    \item \textbf{DA+TM+FBTR.} Same as DA+TM, plus FBTR. Results with the simplified FBTR version (S\_FBTR) are also shown.
    \item \textbf{DA+TM+FBTR+CM.} Same as DA+TM+FBTR, plus correction capabilities.
\end{itemize}

\begin{figure}[t!]
    \centering
    \includegraphics[width=0.95\linewidth]{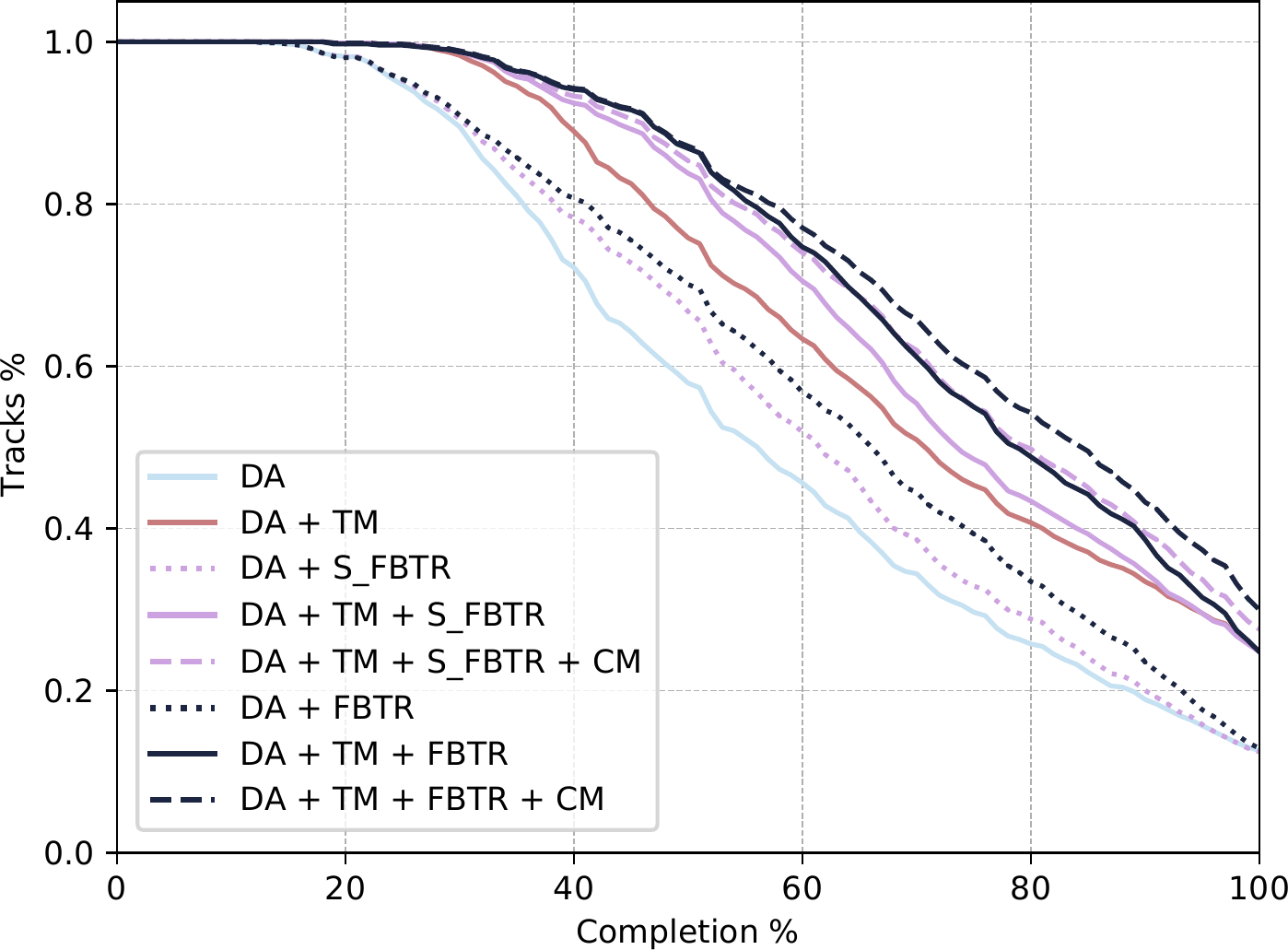}
    \caption{CRPs for the different ablation experiments.}
    \label{fig:ablation_CRP}
\end{figure}

Table \ref{tab:ablation_ALL_table} and Figure \ref{fig:ablation_CRP} show the results of the ablation study on our whole dataset. It can be clearly observed that each added module increases the overall tracking completion rate. 

The impact of FBTR is particularly noteworthy: it increases the CRS by 9.8\% (from 0.593 to 0.651) and 7.3\% (from 0.710 to 0.762) when added to DA and DA+TM, respectively. The rank-based FBTR constraint demonstrates its efficacy by enhancing the CRS of its simplified version by 4.3\% (from 0.624 to 0.651) and 3.3\% (from 0.738 to 0.762) when added to DA and DA+TM, respectively.
The correction module also improves long-term tracking (CRS increase of 3.4\% with S\_FBTR, from 0.738 to 0.763, and 2.8\% with FBTR, from 0.762 to 0.783) without any extra computational cost. At the same time, it strongly reduces the fragmentation generated by the FBTR module by 19.1\% (from 0.1850 to 0.1496) and 26.6\% (from 0.1896 to 0.1391), for both simplified and rank-based FBTR, respectively. As expected, the number of ID-Switches increases as we achieve longer-term tracking, but its value stays reasonably low (0.00497 and 0.00530 for S\_FBTR and FBTR, respectively) and slightly decreases when using the CM module (0.00486 and 0.00512, respectively).

Table \ref{tab:ablation_tables} details ablation results per video. Results on \textit{Choke1} and \textit{Choke2} highlight the good performance of our architecture, especially of the FBTR module (CRS increases above 60\%), in contexts where people leave and re-enter the scene. The impact of the CM is also dramatic, reaching CRS values up to 0.940 and reducing fragmentation up to 74\%.
The remaining videos do not contain subject re-appearances. In the case of  \textit{Sidewalk} and \textit{Bengal}, where long occlusions are frequent, FBTR improves short-term tracking by 19.0\% (CRS$=$0.796) and 15.3\% (CRS$=$0.844), respectively. In \textit{Street}, where camera angle is lower, the FBTR module becomes more vulnerable to ID-switches (24.9\% of increase), even though it still considerably increases the CRS (3.6\%). \textit{Terminal 1, 2, 3} and \textit{4} videos are by far the most challenging in terms of illumination, motion and occlusions. In these videos, the impact of FBTR and CM is lower, but the DA+TM+FBTR+CM architecture is still the most successful one. In \textit{Shibuya}, which provides a much higher resolution (3840$\times$2160), the face detector becomes an important bottleneck, propagating a dramatic speed decrease in all combinations (up to 3.3 FPS). At fixed resolutions (1080p), we can appreciate a higher decrease on computational performance for the FBTR module in videos with higher density of faces, such as \textit{Sidewalk} (from 15.5 to 5.4 FPS) or \textit{Bengal} (from 19.9 to 8.5 FPS).


\begin{table}[t!]
    \centering
    \small
    \begin{tabular}{lcccc}
        \toprule
        Tracker & Frags & ID-Switches & CRS & FPS\\
        \midrule
        DA & 0.04302 & \textbf{0.00161} & 0.593 & \textbf{36.2}\\
        DA+TM & 0.01650 & 0.00469 & 0.710 & 26.4\\
        \midrule
        DA+S\_FBTR & 0.04569 & 0.00172 & 0.624 & 13.3\\
        DA+TM+S\_FBTR & 0.01850 & 0.00497 & 0.738 & 11.7\\
        DA+TM+S\_FBTR+CM & 0.01496 & 0.00486 & 0.763 & 11.7\\
        \midrule
        DA+FBTR & 0.04613 & 0.00214 & 0.651 & 13.2\\
        DA+TM+FBTR & 0.01896 & 0.00530 & 0.762 & 11.7\\
        DA+TM+FBTR+CM & \textbf{0.01391} & 0.00512 & \textbf{0.783} & 11.7\\
        \bottomrule
    \end{tabular}
    \caption{Results of the ablation study on our whole dataset.}
    \label{tab:ablation_ALL_table}
\end{table}

\subsection{Distractor analysis}

We define a distractor as a person (identity) that does not appear in the set of videos used for validation, but that is registered in the system to cause potential failures (misidentifications) in the face verification procedure. Deploying the proposed method in real surveillance scenarios would imply considerably longer videos and continuous video stream analysis of crowded scenes, which would result in a high number of inactive tracklets from previously seen identities. In this experiment, we analyze how our system would perform in such scenarios, in which the number of past observed identities substantially increases. To do so, we create ``ghost tracklets'': a set of tracklets made of enrollable and verifiable faces from distractors. Ghost tracklets are registered in the FBTR module before the video processing starts. Then, we re-run the complete architecture without the CM extension. 

Table~\ref{tab:distractors_analysis} shows the metrics relative to three distractor sets of increasing size: 200, 2K and 20K distractors. Although the computational performance drops with the number of distractors (3\%, 23\%, and 71\%, respectively), the resulting values for fragmentation, ID-switches and CRS remain stable. These results anticipate the effectiveness of the presented method when applied to longer videos with higher number of identities.

\begin{table}[ht!]
    \centering
    \small
    \begin{tabular}{lccccc}
        \toprule
        Choke1 (800x600) & Frag & ID-Switches & CRS & FPS\\
        \midrule
            DA & 0.01607 & \textbf{0.00001} & 0.375 & \textbf{87.0}\\
            DA+TM & 0.01017 & 0.00013 & 0.397 & 61.7\\
            DA+TM+FBTR & 0.01745 & 0.00025 & 0.823 & 22.8\\
            DA+TM+FBTR+CM & \textbf{0.00439} & 0.00038 & \textbf{0.940} & 22.8\\
        \midrule
        
        \multicolumn{5}{l}{Choke2 (800x600)}\\
        \midrule
            DA & 0.03387 &\textbf{ 0.00069} & 0.355 & \textbf{87.3}\\
            DA+TM & 0.01584 & 0.00253 & 0.400 & 55.9\\
            DA+TM+FBTR & 0.02181 & 0.00230 & 0.674 & 18.9\\
            DA+TM+FBTR+CM & \textbf{0.01412} & 0.00253 & \textbf{0.872} & 18.9\\
        \midrule
        
        \multicolumn{5}{l}{Terminal1 (1920x1080)}\\
        \midrule
            DA & 0.05196 & \textbf{0.00197} & 0.602 & \textbf{26.1}\\
            DA+TM & 0.01530 & 0.00466 & 0.741 & 20.2\\
            DA+TM+FBTR & 0.01596 & 0.00488 & 0.758 & 10.5\\
            DA+TM+FBTR+CM & \textbf{0.01450} & 0.00481 & \textbf{0.763} & 10.5\\
        \midrule
        
        \multicolumn{5}{l}{Terminal2 (1920x1080)}\\
        \midrule
            DA & 0.06025 & \textbf{0.00208} & 0.649 & \textbf{26.3}\\
            DA+TM & 0.01654 & 0.00701 & 0.789 & 21.0\\
            DA+TM+FBTR & 0.01697 & 0.00753 & 0.794 & 11.1\\
            DA+TM+FBTR+CM & \textbf{0.01567} & 0.00745 & \textbf{0.801} & 11.1\\
        \midrule
        
        \multicolumn{5}{l}{Terminal3 (1920x1080)}\\
        \midrule
            DA & 0.05405 &\textbf{ 0.00259} & 0.613 & \textbf{26.3}\\
            DA+TM & 0.01974 & 0.00423 & 0.774 & 21.1\\
            DA+TM+FBTR & 0.02092 & 0.00423 & 0.789 & 11.0\\
            DA+TM+FBTR+CM & \textbf{0.01810} & 0.00423 &\textbf{ 0.791} & 11.0\\
        \midrule
        
        \multicolumn{5}{l}{Terminal4 (1920x1080)}\\
        \midrule
            DA & 0.07031 & \textbf{0.00326} & 0.577 & \textbf{25.8}\\
            DA+TM & 0.03360 & 0.00918 & 0.668 & 19.5\\
            DA+TM+FBTR & 0.03493 & 0.01007 & 0.694 & 9.7\\
            DA+TM+FBTR+CM & \textbf{0.03049} & 0.00933 & \textbf{0.702} & 9.7\\
        \midrule
        
        \multicolumn{5}{l}{Sidewalk (1920x1080)}\\
        \midrule
            DA & 0.02929 & \textbf{0.00237} & 0.547 & \textbf{24.3}\\
            DA+TM & 0.00996 & 0.00984 & 0.669 & 15.5\\
            DA+TM+FBTR & 0.01293 & 0.00996 & 0.796 & 5.4\\
            DA+TM+FBTR+CM & \textbf{0.00759} & 0.00913 & \textbf{0.826} & 5.4\\
        \toprule
        
        \multicolumn{5}{l}{Bengal (1920x1080)}\\
        \midrule
            DA & 0.02459 & \textbf{0.00058} & 0.587 & \textbf{25.6}\\
            DA+TM & 0.01222 & 0.00288 & 0.732 & 19.9\\
            DA+TM+FBTR & 0.01711 & 0.00302 & 0.844 & 8.5\\
            DA+TM+FBTR+CM & \textbf{0.00863} & 0.00273 & \textbf{0.863} & 8.5\\
        \midrule
        
        \multicolumn{5}{l}{Street (1920x1080)}\\
        \midrule
            DA & 0.03031 & \textbf{0.00123} & 0.580 & \textbf{26.6}\\
            DA+TM & 0.01884 & 0.00410 & 0.634 & 23.5\\
            DA+TM+FBTR & 0.02048 & 0.00512 & 0.657 & 15.8\\
            DA+TM+FBTR+CM & \textbf{0.01556} & 0.00512 & \textbf{0.694} & 15.8\\
        \midrule
        
        \multicolumn{5}{l}{Shibuya (3840x2160)}\\
        \midrule
            DA & 0.05038 & \textbf{0.00112} & 0.644 & \textbf{6.2}\\
            DA+TM & 0.01688 & 0.00372 & 0.765 & 5.7\\
            DA+TM+FBTR & 0.01787 & 0.00484 & 0.798 & 3.3\\
            DA+TM+FBTR+CM & \textbf{0.01365} & 0.00459 & \textbf{0.811} & 3.3\\
        \bottomrule
    \end{tabular}
    \caption{Contribution of the proposed modules to each video.}
    \label{tab:ablation_tables}
\end{table}

\begin{table}[t!]
    \centering
    \small
    \begin{tabular}{rccccc}
        \toprule
        Distractors & Frags & ID-Switches & CRS & FPS\\
        \midrule
             0 & 0.01896 & 0.00530 & 0.762 & 11.7\\
             200 & 0.01895 & 0.00536 & 0.761 & 11.4\\
             2000 & 0.01895 & 0.00534 & 0.761 & 9.1\\
             20000 & 0.01896 & 0.00524 & 0.761 & 3.4\\
        \bottomrule
    \end{tabular}
    \caption{Distractors analysis results.}
    \label{tab:distractors_analysis}
\end{table}

%% file: sections/06_conclusions.tex
\section{Conclusions}

We have presented an architecture for long-term, multi-face tracking in crowded video-surveillance scenarios. The proposed method benefits from advances in the fields of face detection and face recognition to achieve long-term tracking, in contexts that are particularly unconstrained in terms of movement, re-appearances, and occlusions. 
We have introduced specialized metrics conceived to evaluate long-term tracking capabilities, and publicly released a dataset with ten videos representing the targeted use case. 

The series of experiments carried out lead to interesting findings. Firstly, we demonstrate that our novel tracklet reconnection strategy, grounded on rank-based face verification, allows us to obtain up to 50\% longer tracks. Secondly, we show how the proposed rank-based constraint helps to keep higher inter-class distances among tracklets, while minimizing the propagation of errors due to mixed-up identities. As a result, the completion rate increases up to a 4.3\%, without significant penalties in terms of ID-Switches or fragmentation. Finally, the proposed cost-free correction module has been proved to increase tracking robustness, not only by keeping on improving long-term capabilities, but also by reducing fragmentation.
